\documentclass[journal]{IEEEtran}

\def\BibTeX{{\rm B\kern-.05em{\sc i\kern-.025em b}\kern-.08emT\kern-.1667em\lower.7ex\hbox{E}\kern-.125emX}}

\usepackage[utf8]{inputenc}
\usepackage[T1]{fontenc}
\usepackage{booktabs}
\usepackage{color, colortbl}
\usepackage{soul}
\soulregister\cite7
\soulregister\ref7
\soulregister\pageref7
\soulregister\url7
\soulregister\textit7
\usepackage{amsmath}
\usepackage{comment}
\usepackage{bm}
\usepackage{amssymb}
\usepackage{balance}
\usepackage{url}

\definecolor{lightyellow}{cmyk}{0,0,0.50,0}
\sethlcolor{lightyellow}
\definecolor{yellow}{cmyk}{0,0,0.50,0}

\ifCLASSINFOpdf
   \usepackage[pdftex]{graphicx}
  \DeclareGraphicsExtensions{.pdf,.jpeg,.png}
\else
\fi

\usepackage[tight,footnotesize]{subfigure}

\usepackage{tikz}
\usetikzlibrary{automata}
\usetikzlibrary{shapes.geometric,positioning}
\usetikzlibrary{matrix}

\newsavebox\verbtestsetuno
\newsavebox\verbtestsetdue

\begin{document}

\title{Adversarial machine learning\\for protecting against online manipulation}

\author{Stefano~Cresci, 
        Marinella~Petrocchi,
        Angelo~Spognardi,
        and~Stefano~Tognazzi
\thanks{S. Cresci is with the Institute of Informatics and Telematics, IIT-CNR, Pisa, Italy.}
\thanks{M. Petrocchi is with the Institute of Informatics and Telematics, IIT-CNR, Pisa, Italy and Scuola IMT Alti Studi Lucca, Lucca, Italy.}
\thanks{A. Spognardi is with the Dept. of Computer Science, Universit\`{a} di Roma `La Sapienza', Roma, Italy.}
\thanks{S. Tognazzi is with the Centre for the Advanced Study of Collective Behaviour and the Dept. of Computer and Information Science, Konstanz University, Konstanz, Germany.}}

\maketitle


\begin{abstract}
Adversarial examples are inputs to a machine learning system that result in an incorrect output from that system. Attacks launched through this type of input can cause severe consequences: for example, in the field of image recognition, a stop signal can be misclassified as a speed limit indication.
However, 
adversarial examples 
also represent the fuel for a flurry of research directions in  different domains and applications. Here, we give an overview of how they can be profitably exploited as powerful tools to build stronger learning models, capable of better-withstanding attacks, for two crucial tasks: fake news and social bot detection.
\end{abstract}


\IEEEpeerreviewmaketitle
\begin{figure*}
  \centering
  \subfigure[\textbf{Computer vision.} Images can be modified by adding adversarial patches so as to fool image classification systems (e.g., those used by autonomous vehicles).\label{fig:adversarial-examples-CV}]{\includegraphics[width=.274\textwidth]{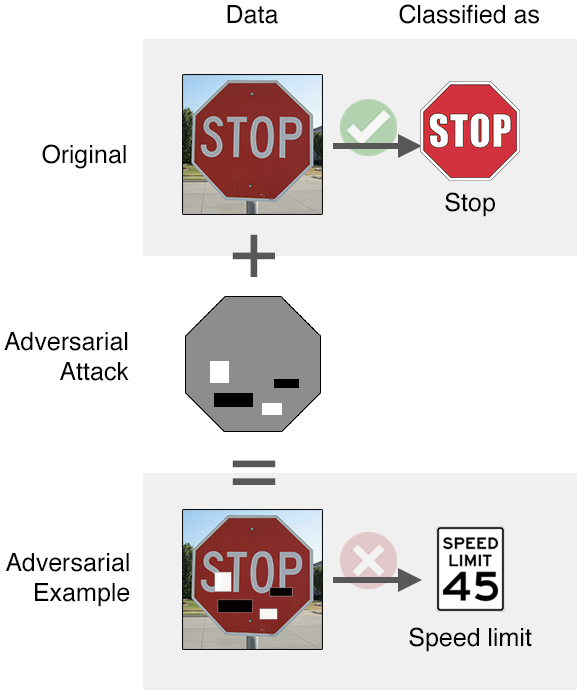}}\hspace{.02\textwidth}
  \subfigure[\textbf{Automatic speech recognition.} Adding adversarial noise to a speech waveform may result in wrong textual translations.\label{fig:adversarial-examples-SR}]{\includegraphics[width=0.205\textwidth]{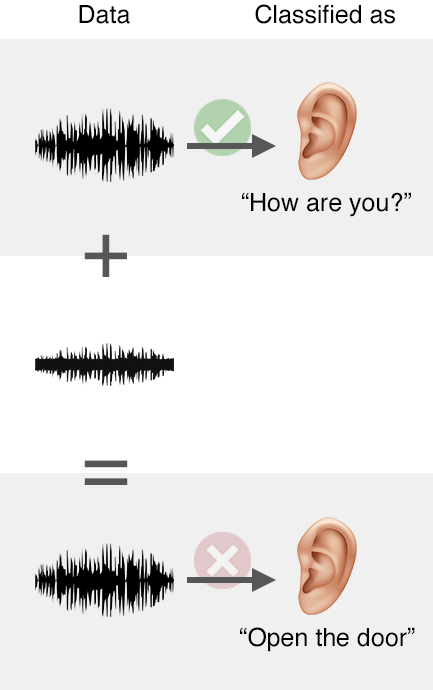}} \hspace{.02\textwidth}
  \subfigure[\textbf{Social bot detection.} Similarly to computer vision and automatic speech recognition, adversarial attacks can alter the features of social bots, without impacting their activity, thus allowing them to evade detection.\label{fig:adversarial-examples-BD}]{\includegraphics[width=0.205\textwidth]{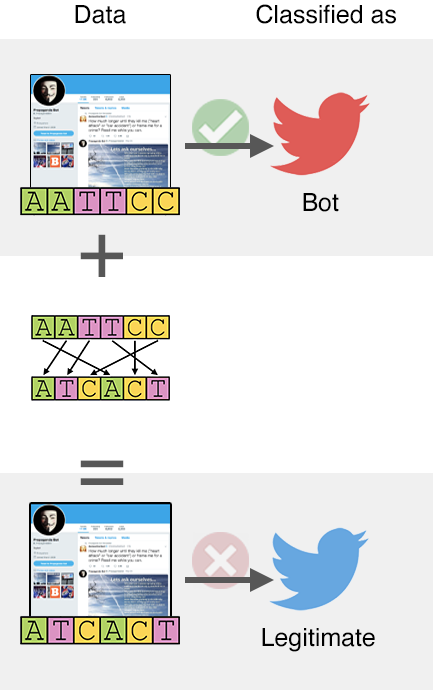}}\hspace{.02\textwidth}
  \subfigure[\textbf{Fake news detection.} Tampering with the textual content of an article, or even with its comments, may yield wrong article classifications.\label{fig:adversarial-examples-FN}]{\includegraphics[width=0.205\textwidth]{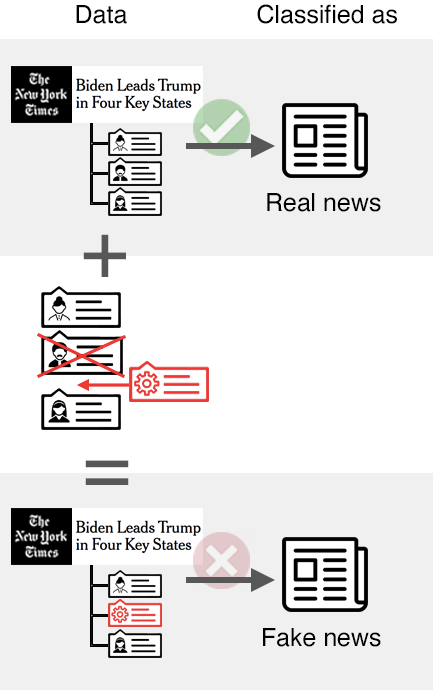}}
  \caption{Adversarial examples and their consequences, for a few notable ML tasks.\label{fig:adversarial-examples}}
\end{figure*}

\section*{}
\label{sec:adversarial}
The year was 1950, and in his paper `Computing Machinery and Intelligence', Alan Turing asked this question to his audience: {\it Can a machine think rationally}?
A question partly answered by the machine learning (ML) paradigm, whose traditional definition is as follows: ``A computer program is said to learn from experience {\it E} with respect to some class of tasks {\it T} and performance measure {\it P}, if its performance at tasks in {\it T}, as measured by {\it P}, improves with experience {\it E}''~\cite{Mitchell1997}. If we define the experience {\it E} as `what data to collect’, the task {\it T} as `what decisions the software needs to make’, and the performance measurement {\it P} as `how we will evaluate its results’, then it becomes possible to evaluate the capability of the program to complete the task correctly -- that is, to recognize the type of data, evaluating its performance.  

To date, ML helps us to achieve multiple goals, it provides recommendations to customers based on their previous purchases or gets rid of spam in the inbox based on spam received previously, just to name a couple of examples. In the image recognition field, the above program has been trained by feeding it different images, thus learning to distinguish them. However, in 2014, Google and NYU researchers showed how it was possible to fool the classifier, specifically an ensemble of neural networks called ConvNets, by adding noise to the image of a panda. The program classified the panda plus the added noise as a gibbon, with a 99\% confidence. The modified image is called {\it adversarial example}~\cite{GoodfellowSS14}. Formally, given a data distribution $p(x,y)$ over images $x$ and labels $y$ and a classifier $f$ such that $f(x)=y$, an adversarial example is a modified input $\tilde{x} = x + \delta$ such that $\delta$ is a very small (human-imperceptible) perturbation and $f(\tilde{x}) \ne y$, namely $\tilde{x}$ is misclassified while $x$ was not. Still in visual recognition, it is possible to `perturb' a road sign that reproduces, e.g., a stop sign by placing small stickers on it, so that the classifier identifies it as a speed limit sign~\cite{Eykholt2018}, see  Figure~\ref{fig:adversarial-examples-CV}. Another noteworthy attack exploits the so-called `adversarial patches': the opponent does not even need to know the target that the classifier has been trained to recognize. Simply adding a patch to the input can lead the system to decide that what it has been given to classify is exactly what the patch represents. It became popular the case of the trained model exchanging a banana for a toaster, having patched an adversarial example next to the banana~\cite{brown2018adversarial}. 
The few previous examples demonstrate the risks caused by the
vulnerability of ML systems to intentional data manipulations, for the field of computer vision. 

In recent years, ML was also at the core of a plethora of efforts in other domains. In particular, some of those
that have seen massive application of ML and AI are intrinsically adversarial -- that is, they feature the natural and inevitable presence of adversaries motivated in fooling the ML systems. In fact, adversarial examples are extremely relevant in all security-sensitive applications, where any misclassification induced by an attacker represents a security threat. A paramount example of such applications is the fight against online abuse and manipulation, which often come under the form of fake news and social bots. 

So, how to defend against attackers who try to deceive the model through adversarial examples? It turned out that adversarial examples are not exclusively a threat to the reliability of ML models. Instead, they can also be leveraged as a very effective mean to strengthen the models themselves. A `brute force' mode, the so-called {\it Adversarial Training}, sees the model designers pretend to be attackers: they generate several  adversarial examples against their own model and, then, train the model not to be fooled by them.

\begin{figure*}
  \centering
  \includegraphics[width=1\textwidth]{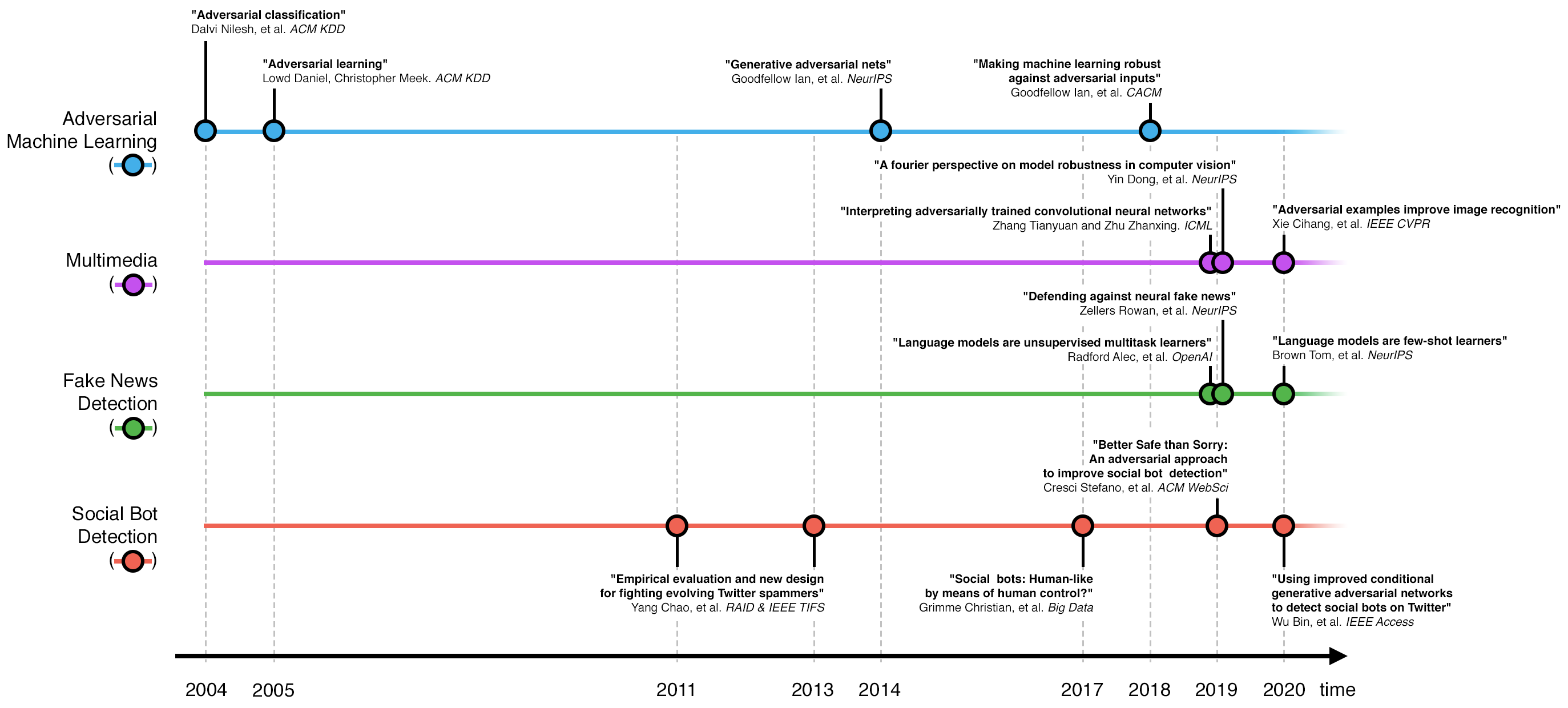}
  \caption{Adversarial machine learning lead to a rise of adversarial approaches for the detection of manipulated multimedia, fake news, and social bots.\label{fig:adversarial-timeline}}
\end{figure*}

Along these lines, in the remainder we briefly survey relevant literature on adversarial examples and {\it Adversarial Machine Learning} (AML). AML aims at understanding when, why, and how learning models can be attacked, and the techniques that can mitigate attacks. We will consider two phenomena whose detection is polluted by adversaries and that are bound to play a crucial role in the coming years for the security of our online ecosystems: fake news and social bots. Contrary to computer vision, the adoption of AML in these fields is still in its infancy, as shown in Figure~\ref{fig:adversarial-timeline}, despite its many advantages. Current endeavors mainly focus on the identification of adversarial attacks, and only seldom on the development of solutions that leverage adversarial examples for improving detection systems. The application of AML in these fields is still largely untapped, and its study will provide valuable insights for driving future research efforts and getting practical advantages.

\section*{Fake news and social bots}\label{sec:fakenews}

Fake news are often defined as `fabricated information that mimics news media content in form but not in organizational process or intent’~\cite{Lazer2018,Quandt2019}. Their presence has been documented in several contexts, such as politics, vaccinations, food habits and financial markets.

False stories have always circulated centuries before the Internet. One thinks, for instance, of the manoeuvres carried out by espionage and counter-espionage to get the wrong information to the enemy. 
If fake stories have always existed, why are we so concerned about them now? 
The advent of the Internet, while undoubtedly facilitating the access to news, has lowered the editorial standards of journalism, and its open nature has led to a proliferation of user-generated content, un-screened by any moderator~\cite{gangware2019weapons}.

Usually, fake news is published on some little-known outlet, and amplified through social media posts, quite often using so-called social bots. Those are software algorithms that can perfectly mimic the behaviour of a genuine account and maliciously generate artificial hype~\cite{cresci2020decade,Xinyi2020survey}.

In recent years, research has intensified efforts to combat both the creation and spread of fake news, as well as the use of social bots.
Currently, the most common detection methods for both social bots (as tools for spreading) and fake news are based on supervised machine learning algorithms. In many cases, these approaches achieve very good performances on considered test cases.
Unfortunately, state-of-the-art detection techniques 
suffer from attacks that critically degrade the performances of the learning algorithms. 
In a classical adversarial game, social bots evolved over time~\cite{cresci2020decade}: while early bots in the late 2000s were easily detectable only by looking at static account information or simple indicators of activity, sophisticated bots are nowadays almost indistinguishable from genuine accounts. 
We can observe the same adversarial game in fake news. Recent studies show that it is possible to subtly act on the title, content, or source of the news, to invert the result of a classifier: from true to false news, and vice versa~\cite{Horne2019newsattacks}.

\section*{Adversarial fake news detection}
Learning algorithms have been adopted with the aim of 
detecting false news by, e.g., using textual features, such as the title 
and content of the article. Also, it has been shown that users' 
comments and replies can be valid features to unveil low or no 
reputable textual content~\cite{Shu2019dEFENDE}.

Regrettably. algorithms can be fooled. As an example, TextBugger~\cite{LiJDLW19} is a general attack framework for generating adversarial text that  can trick sentiment analysis classifiers, such as \cite{convolutionalltexxtclassification}, into erroneous classifications via marginal modifications of the text, such as adding or removing individual words, or even single characters. 
Moreover, not only can a fake news classifier be fooled by tampering with part of the news, but also by acting on comments and replies.
Figure~\ref{fig:adversarial-examples-FN} exemplifies the attack: a
detector correctly identifies a real article as indeed real. Unfortunately, 
by inserting a fake comment as part of its inputs, the same detector is 
misled to predict the article as fake instead. Fooling fake news 
detectors via adversarial comment generation has been demonstrated 
feasible by Le et al. in~\cite{le2020malcom}. Leveraging the alteration of social responses,  
such as comments and replies, to fool the classifier prediction is 
advantageous because the attacker does not have to own the published 
piece (in order to be able to modify it after the publication), and the 
passage from `written-by-humans’ to self-generated text is less 
susceptible to detection by the naked eye. In fact, comments and replies 
are usually accepted, even if written in an informal style and with scarce 
quality.
Also, work in~\cite{le2020malcom} shows how it is possible to 
generate adversarial comments of high quality and relevance with the 
original news, even at the level of the whole sentence.


Recent advances in text generation make even possible to generate coherent paragraphs of text. This is the case, for example, of GPT-2~\cite{radford2019language}, a language model trained on a dataset of 8M Web pages. 
Being trained on a myriad of different subjects, GPT-2 leads to the generation of surprisingly high quality texts, outperforming other language models with domain-specific training (like news, or books). Furthermore, in~\cite{Mosallanezhad2020topic}, the authors study how to preserve a topic in synthetic news generation. Contrary to GPT-2, which selects the most probable word from the vocabulary as the next word to generate, a reinforcement learning agent tries to select words that optimize the matching of a given topic.

Achievements in text generation have positive practical implications, such as, e.g., translation. Concerns, however, have arisen because malicious actors can exploit these generators to produce false news automatically. While  most  of  online disinformation  today  is  manually  written, as  progress  continues in  natural  language  text  generation,  the creation of propaganda and realistic-looking hoaxes   will grow at scale~\cite{dasanmartino2020survey}.
In \cite{NEURIPS2019}, for example, the authors present Grover, a model for controllable text generation, with the aim of defending against fake news.  Given a headline, Grover can generate the rest of the article, and vice versa. Interestingly, investigating the level of credibility, articles generated with propagandist tone result more credible to human readers, rather than articles with the same tone, but written by humans. 
If, on the one hand, this shows how to exploit text generators to obtain `reliable fake news', on the other hand, it is the double-edged blade that allows the reinforcement of the model. Quoting from~\cite{NEURIPS2019}: `the best defense against Grover turns out to be Grover itself', as able to achieve 92\% accuracy in discriminating between human-written and auto-generated texts. 
Grover is just one of other news generators that obtain noticeable results: a vast majority of the news generated by Grover and four others can fool human readers, as well as a neural network classifier, specifically trained to detect fake news~\cite{Mosallanezhad2020topic}. Texts generated by the recent upgrade of GPT-2, GPT-3, get even more impressive results in resembling hand-written stories~\cite{brown2020language}.

Finally, Miller et al. consider the discrimination between true and fake news  very challenging~\cite{Miller2020adversarial}: it is enough, e.g., to change a verb from the positive to the negative form to completely change the meaning of the sentence. They therefore see the study of the news source as a possible way of combating this type of attack.

\section*{Adversarial social bot detection}
The roots of adversarial bot detection date back to 2011 and almost coincide with the initial studies on bots themselves~\cite{cresci2020decade}.
Between 2011 and 2013 -- that is, soon after the first efforts for detecting automated online accounts -- several scholars became aware of the \textit{evolutionary} nature of social bots. In fact, while the first social bots that inhabited our online ecosystems around 2010 were extremely simple and visibly untrustworthy accounts, those that emerged in subsequent years featured increased sophistication. This change was the result of the development efforts put in place by botmasters and puppeteers for creating automated accounts capable of evading early detection techniques~\cite{cresci2017paradigm}. Comparative studies between the first bots and subsequent ones, such as those in~\cite{Yang:2013}, unveiled the evolutionary nature of social bots and laid the foundations for adversarial bot detection. Notably, bot evolution still goes on, fueled by the latest advances in powerful computational techniques that allow mimicking human behavior better than ever before~\cite{boneh2019relevant}.

Based on these initial findings, since 2011 some initial solutions were proposed for detecting evolving social bots~\cite{Yang:2013}.
These techniques, however, were still based on traditional approaches to the task of social bot detection, such as those based on general purpose, supervised machine learning algorithms~\cite{cresci2020decade}. Regarding the methodological approach, the novelty of this body of work mainly revolved around the identification of those machine learning features that seemed capable of allowing the detection of the sophisticated bots. The test of time, however, proved such assumptions wrong. In fact, those features that initially seemed capable of identifying the sophisticated bots, started yielding unsatisfactory performance soon after their proposal~\cite{cresci2017paradigm}.

It was not until 2017 that adversarial social bot detection really ignited. Since then, several approaches were proposed in rapid succession for testing the detection capabilities of existing bot detectors, when faced with artfully created adversarial examples. Among the first adversarial examples of social bots there were accounts that did not exist yet, but whose behaviors and characteristics were simulated, as done in~\cite{CresciOSNEM2019,Cresci2019WebSci}. There, authors used genetic algorithms to `optimize' the sequence of actions of groups of bots so that they could achieve their malicious goals, while being largely misclassified as legitimate, human-operated accounts. Similarly,~\cite{he2021petgen} trained a text-generation deep learning model based on latent user representations (i.e., embeddings) to create adversarial fake posts that would allow malicious users to escape Facebook's detector TIES~\cite{noorshams2020ties}. Other adversarial social bot examples were accounts developed and operated ad-hoc for the sake of evaluating the detection capabilities of existing bot detectors, as done in~\cite{grimme2017social}. 
Experimentation with such examples helped scholars understand the weaknesses of existing bot detection systems, as a first step for improving them. However, the aforementioned early body of work on adversarial social bot examples still suffered from a major drawback. All such works adopted ad-hoc solutions for generating artificial bots, thus lacking broad applicability. Indeed, some solutions were tailored for testing specific detectors~\cite{CresciOSNEM2019,Cresci2019WebSci}, while others relied on manual interventions, thus lacking scalability and generality~\cite{grimme2017social}.

With the widespread recognition of AML as an extremely powerful learning paradigm, also came new and state-of-the-art approaches for adversarial social bot detection. A paramount example of this spillover is the work proposed in~\cite{wu2020using}. There, authors leverage a generative adversarial network (GAN) for artificially generating a large number of adversarial bot examples with which they trained downstream bot detectors.
Results demonstrated that this approach augments the training phase of the bot detector, thus significantly boosting its detection performance. Similarly, a GAN is also used in~\cite{zheng2019one} to generate latent representations of malicious users solely based on the representations of benign ones. The representations of the real benign users are leveraged in combination with the artificial representations of the malicious users to train a discriminator for distinguishing between benign and malicious users.

\section*{Conclusions}
\label{sec:conclusions}
The success of a learning system is crucial in many scenarios of our life, be it virtual or not: correctly recognizing a road sign, or discriminating between genuine and fake news. Here, we focused on adversarial examples---created to fool a trained model, and on AML---which exploits such examples to strengthen the model.

Remarkably, adversarial examples, originally in the limelight especially in the field of computer vision, now threaten various domains. We concentrated on the recognition of false news and false accounts and we highlighted how, despite the antagonistic nature of the examples, scholars are moving proactively to let attack patterns be curative and reinforce the learning machines. Outside computer vision, these efforts are still few and far between. Improvements along this direction are thus much needed, especially in those domains that are naturally polluted by adversaries.

\ifCLASSOPTIONcaptionsoff
  \newpage
\fi

\balance
\vfill
\end{document}